\documentclass{article}
\usepackage{spconf,amsmath,graphicx,bm,amssymb}
\usepackage{algorithm}
\usepackage{algorithmic}
\usepackage{booktabs}
\usepackage{mathtools}
\usepackage{cite}
\usepackage[hidelinks]{hyperref}
\usepackage{xcolor}
\newcommand{\rev}[1]{{\color{black}#1}}

\DeclareMathOperator{\grad}{grad}
\DeclareMathOperator{\prox}{prox}
\newcommand{\R}{\mathbb{R}}
\newcommand{\ba}{\bm{a}}
\newcommand{\bA}{\bm{A}}
\newcommand{\bD}{\bm{D}}
\newcommand{\bI}{\bm{I}}
\newcommand{\bU}{\bm{U}}
\newcommand{\bV}{\bm{V}}
\newcommand{\bW}{\bm{W}}
\newcommand{\bX}{\bm{X}}
\newcommand{\bY}{\bm{Y}}
\newcommand{\bZ}{\bm{Z}}
\newcommand{\bxi}{\bm{\xi}}
\newcommand{\bzero}{\bm{0}}
\newcommand{\one}{\mathbf{1}}
\newcommand{\Fcal}{\mathcal{F}}
\newcommand{\Mcal}{\mathcal{M}}
\newcommand{\Xcal}{\mathcal{X}}
\newcommand{\Ycal}{\mathcal{Y}}
\newcommand{\Proj}{\operatorname{Proj}}
\newcommand{\Retr}{\operatorname{Retr}}
\newcommand{\dist}{\operatorname{dist}}
\newcommand{\norm}[1]{\left\lVert#1\right\rVert}
\newcommand{\ip}[2]{\langle #1, #2\rangle}
\newsavebox\CBox
\def\textBF#1{\sbox\CBox{#1}\resizebox{\wd\CBox}{\ht\CBox}{\textbf{#1}}}

\newtheorem{definition}{Definition}
\newtheorem{theorem}{Theorem}
\newtheorem{assumption}{Assumption}

\title{Riemannian Difference-of-Convex Optimization for K-Means Clustering}

\name{Meng Xu$^{\star}$, Bo Jiang$^{\dag}$, Hanfu Zhang$^{\ddag}$,
Ya-Feng Liu$^{\ddag}$, and Anthony Man-Cho So$^{\sharp}$}
\address{\parbox{0.96\textwidth}{\centering\normalsize
$^{\star}$AMSS, Chinese Academy of Sciences, and
University of Chinese Academy of Sciences, Beijing, China\\
$^{\dag}$Ministry of Education Key Laboratory of NSLSCS, School of Mathematical Sciences, Nanjing Normal University, Nanjing, China\\
$^{\ddag}$Ministry of Education Key Laboratory of Mathematics and Information Networks,\\
School of Mathematical Sciences, Beijing University of Posts and Telecommunications, Beijing, China\\
$^{\sharp}$Department of Systems Engineering and Engineering Management, The Chinese University of Hong Kong, HKSAR, China
}}

\begin{document}
\ninept
\maketitle

\begin{abstract}
{K-means is a widely adopted clustering approach in signal processing and machine learning.}
In this paper, we study K-means clustering through {a cardinality-constrained} formulation on a compact embedded submanifold. We replace the cardinality constraint with a difference-of-convex (DC) penalty and establish a global error bound to prove that the {penalized and  constrained formulations} share the same global minimizers whenever the penalty parameter exceeds a finite threshold. To {solve} the resulting nonsmooth Riemannian DC problem, we reformulate it as a minimax problem and propose {RADA-DC, a} Riemannian alternating descent ascent method {combining dual regularization with DC linearization.} Under standard assumptions {and suitable parameter choices}, {RADA-DC} finds an \(\epsilon\)-Riemannian critical point within \(\mathcal{O}(\epsilon^{-3})\) iterations. {We conduct experiments} on synthetic and real-world datasets {to} demonstrate that the proposed method outperforms the tested baselines, including K-means++, in solution quality at competitive computational cost when the number of clusters is large.
\end{abstract}


\section{Introduction}
{K-means is one of the most widely used clustering methods in signal processing and machine learning \cite{forero2012robust,jain2010data}.
It partitions data points into a prescribed number of groups by minimizing the sum of squared
distances from each data point to its assigned cluster centroid. Finding a globally optimal {partition}
is NP-hard \cite{aloise2009np}. In practice, Lloyd's algorithm is computationally inexpensive but
sensitive to initialization \cite{lloyd1982least}{; K-means++ mitigates this sensitivity through a carefully designed randomized seeding strategy \cite{arthur2007kmeans}. Obtaining high-quality solutions at a reasonable computational cost remains challenging, particularly when the number of clusters is large \cite{chen2019bigdata}.}
{{This motivates alternative} optimization approaches} based on semidefinite programming (SDP) and linear programming relaxations \cite{peng2007approximating,chen2021cutoff,derosa2026power}. The work
\cite{prasad2018improved} further derived an equivalent generalized completely positive formulation
and, based on it, developed a strengthened SDP relaxation. However, solving these relaxations can be
computationally demanding for large datasets.}

Recent works exploit the geometry of K-means matrix formulations through Riemannian optimization.
In particular, the work \cite{carson2017manifold} considered an exact
formulation over nonnegative matrices with orthonormal columns whose span contains the all-ones vector. It applied a quadratic penalty to nonnegativity while keeping the equality constraints, which define a compact embedded submanifold. Exploiting the invariance of the objective under column sign changes, the work \cite{huang2025riemannian} obtained an equivalent formulation by replacing nonnegativity with a cardinality {constraint and} then used an $\ell_1$ regularizer to promote sparsity on the same manifold. 
Other approaches use nonnegative low-rank models derived from SDP relaxations \cite{zhuang2024statistically,xu2026scalable}. 
These models replace column orthogonality with a Frobenius-norm constraint and allow the factor matrix to have more columns than the number of clusters.
However, the quadratic and \(\ell_1\) penalties in \cite{carson2017manifold,huang2025riemannian} do not generally preserve equivalence to K-means.
We therefore revisit the cardinality-constrained Riemannian formulation and seek a continuous penalty model that is globally exact for a sufficiently large but finite penalty parameter.

{Such finite-parameter equivalence can be established by combining an appropriate error bound for the
feasible set with Lipschitz continuity of the objective on the manifold. Recent works have studied
error bounds and exact penalization for nonnegative orthogonality constraints \cite{jiang2023exact},
sign-constrained Stiefel manifolds \cite{chen2025tight}, extreme point pursuit
\cite{liu2024extreme1,liu2024extreme2}, and sparsity-constrained optimization
\cite{gotoh2018dc,zhao2026exact}. Related penalty formulations have also been considered for
clustering \cite{wang2021clustering,yin2026error}. However, the available error bounds depend on the
specific structure of the constraint sets considered in those works and do not directly yield a
bound for the set defined jointly by the cardinality and manifold constraints in our model.
Motivated by these developments, {\emph{we establish a global error bound that uses a difference-of-convex
(DC) penalty residual to control the distance from any point on the manifold to this joint feasible
set.}} This bound ensures that, above a finite penalty threshold, the Riemannian DC model shares the same global minimizers {as the cardinality-constrained formulation of K-means and is therefore globally equivalent to the original K-means problem.}
}

{A number of algorithms that exploit both the manifold and DC structures have been proposed to solve Riemannian DC optimization problems \cite{bergmann2024difference,jiang2025irpdc,li2024proximal}. Specifically, the} work \cite{bergmann2024difference} developed a Riemannian DCA on Hadamard manifolds, but its Hadamard assumption is not satisfied by the manifold in our K-means formulation. {The works \cite{li2024proximal,jiang2025irpdc}} developed proximal methods for structured nonsmooth optimization over embedded Riemannian submanifolds. However, their proximal subproblems generally require iterative solution at each outer iteration, leading to a double-loop structure. {By} contrast, single-loop methods for nonsmooth Riemannian composite optimization offer simple updates with the best-known iteration complexity guarantees \cite{beck2023dynamic,xu2024riemannian}. These advantages motivate us to develop an efficient single-loop algorithm for nonsmooth Riemannian DC optimization.

{\emph{The main contributions of this paper are an exact Riemannian DC reformulation of the K-means problem based on a new error bound and an efficient algorithm for solving the resulting model.}} Building on the Riemannian alternating descent ascent (RADA) framework \cite{xu2024riemannian}, {\emph{we propose RADA-DC for general nonsmooth Riemannian DC optimization.}} The {proposed} method combines RADA's dual regularization with the {standard DC linearization strategy \cite{lethi2018dc}} to handle the additional nonsmooth concave term. Each outer iteration performs a fixed number of Riemannian gradient descent steps followed by a proximal dual update. Under standard assumptions and suitable parameter choices, we prove that RADA-DC finds an $\epsilon$-Riemannian critical ($\epsilon$-RC) point within $\mathcal{O}(\epsilon^{-3})$  iterations. Our experiments further show that Lloyd's algorithm initialized by RADA-DC attains the lowest K-means objective value in most tested settings with a large number of clusters, at competitive computational cost.

\section{K-Means and Its Riemannian DC Formulation}
In this section, we first recall a {cardinality}-constrained Riemannian
formulation of K-means and then establish its exact Riemannian DC
reformulation through a global error bound.
\subsection{K-Means Formulation}
Let \(\bA=[\ba_1,{\ba_2,}\ldots,\ba_n]^\top\in\R^{n\times d}\) be the data matrix, where \(\ba_i\in\R^d\) denotes the \(i\)-th sample, and let \(K\) be the prescribed number of clusters. We use \(\bI_K\) and \({\one_m\in\R^m}\) to denote  the \(K\times K\) identity matrix and the {\(m\)-dimensional} all-ones vector, respectively.
Throughout this paper,
$\ip{\cdot}{\cdot}$ {and $\norm{\cdot}$ denote} the standard Euclidean inner product and its induced norm, respectively. For a partition
$\{\mathcal C_1,{\mathcal C_2,}\ldots,\mathcal C_K\}$ of $[n]:=\{1,{2,}\ldots,n\}$ into $K$ nonempty clusters, define the
normalized cluster indicator matrix $\bX\in\R^{n\times K}$ by {\cite{boutsidis2009unsupervised}}
\begin{equation}\label{eq:indicator-matrix}
 \bX_{ij}=\begin{cases}
 |\mathcal C_j|^{-1/2}, & \text{if}~i \in\mathcal C_j,\\
 0, & {\text{otherwise}}, %
 \end{cases}
\end{equation}
{where $|\mathcal C_j|$ denotes the number of samples in cluster $\mathcal C_j$.}
{By construction, $\bX$ has orthonormal columns (i.e., $\bX^\top\bX=\bI_K$) and satisfies $\bX\bX^\top{\one_n}={\one_n}$. Moreover, each row of $\bX$ contains exactly one nonzero entry, which is positive.}
{Thus, $\norm{\bX}_0=n$, where}
$\norm{\bX}_0$ {denotes} the number of nonzero entries of $\bX$. Using this
matrix representation, {the work} \cite{huang2025riemannian}
reformulated K-means as the following $\ell_0$-constrained Riemannian
optimization problem:
\begin{equation}\label{eq:kmeans-original}
 \begin{array}{cl}
 {\displaystyle\min_{\bX\in\R^{n\times K}}}
 &f_{\rm km}(\bX):=-\ip{\bA\bA^\top}{\bX\bX^\top}\\
 {\mathrm{s.t.}}
 &\bX^\top\bX=\bI_K,\,\, \bX\bX^\top{\one_n}={\one_n},\,\,  \|\bX\|_0=n.
 \end{array}
\end{equation}
{Although {problem}~\eqref{eq:kmeans-original} does not explicitly impose nonnegativity {constraints}, the work \cite{huang2025riemannian} showed that every feasible $\bX$ is a normalized cluster indicator matrix of the form \eqref{eq:indicator-matrix} up to column-sign changes. Such changes preserve $\bX\bX^\top$ and hence the objective value, ensuring the equivalence. {Let $\Fcal$ be the set defined by the}} first two constraints in \eqref{eq:kmeans-original}{, i.e.,}
\begin{equation}\label{eq:kmeans-manifold}
 {\Fcal}\mathrel{{=}}\{\bX\in\R^{n\times K}{\mid}\bX^\top\bX=\bI_K,
                       \ \bX\bX^\top{\one_n}={\one_n}\},
\end{equation}
which is a compact embedded submanifold of $\R^{n\times K}$
\cite{huang2025riemannian}.
{Since $\bX\bX^\top{\one_n}={\one_n}$ rules out zero rows, the remaining constraint $\norm{\bX}_0=n$ enforces exactly one nonzero entry per row, corresponding to assigning each sample to exactly one cluster.}

\subsection{DC {Penalty} Reformulation}
To tackle the cardinality constraint in \eqref{eq:kmeans-original}, {the work \cite{huang2025riemannian} adopted the widely used $\ell_1$-penalty approach to promote sparsity while retaining the manifold constraint $\bX\in\Fcal$. However, this regularization does not guarantee that the resulting model shares the same global minimizers as \eqref{eq:kmeans-original}. We therefore seek a DC penalty that preserves these minimizers for a sufficiently large but finite penalty parameter.}

Let $\norm{\bX}_1\mathrel{{=}}\sum_{i=1}^n\sum_{j=1}^K|\bX_{ij}|$ be the entrywise
$\ell_1$ norm {of $\bX$.} Let
$|\bX|_{[1]}\mathrel{\rev{\geq}}|\bX|_{[2]}\geq\cdots\geq|\bX|_{[nK]}$ be the absolute entries of $\bX$
arranged in nonincreasing order, and {denote} the largest-$n$ norm by
$\norm{\bX}_{[n]}\mathrel{{=}}\sum_{i=1}^n|\bX|_{[i]}$.
Following the work \cite{gotoh2018dc}, the DC residual $r_{\rm DC}(\bX)\coloneqq\norm{\bX}_1-\norm{\bX}_{[n]}$ vanishes if and only if $\norm{\bX}_0\leq n$. Let $\Xcal\coloneqq\{\bX\in\Fcal\mid\norm{\bX}_0=n\}$ {be} the feasible set of \eqref{eq:kmeans-original}. Since every $\bX\in\Fcal$ has at least $n$ nonzero entries, we have
\begin{equation}\label{eq:dc-constraint-function}
 {\bX\in\Xcal\quad\Longleftrightarrow\quad\bX\in\Fcal\quad{\text{and}}\quad r_{\rm DC}(\bX)=0.}
\end{equation}
{Moreover,} both $\norm{\cdot}_1$ and $\norm{\cdot}_{[n]}$ are convex {and their difference $r_{\rm DC}(\bX)$ is nonnegative}. {Thus,} penalizing
the equality in \eqref{eq:dc-constraint-function} yields the following
Riemannian DC {penalty formulation for K-means:}
\begin{equation}\label{eq:penalized-dc}
 \min_{\bX\in{\Fcal}}
 f_{\rm km}(\bX)+\tau\big(\|\bX\|_1-\|\bX\|_{[n]}\big),
\end{equation}
where $\tau>0$ is a sparsity penalty parameter.  
To establish its equivalence to \eqref{eq:kmeans-original} for a finite
{$\tau$}, we first {derive} the following global error bound.
\begin{theorem}[Global error bound]\label{thm:error-bound}
{Let $n\geq K\geq2$. Then, for any $\bX\in\Fcal$, we have}
\begin{equation}\label{eq:error-bound}
 \dist(\bX,\Xcal)\leq 8K\sqrt{nK}\,r_{\rm DC}(\bX),
\end{equation}
where $\dist(\bX,{\Xcal})$ denotes the distance from $\bX$ to {$\Xcal$.}
\end{theorem}

{Since $f_{\rm km}$ is {$\sqrt K\norm{\bA}_2^2$-Lipschitz} continuous on $\Fcal$, where $\norm{\bA}_2$ denotes the spectral norm of $\bA$, Theorem~\ref{thm:error-bound} and \cite[Lemma~5]{liu2024extreme1} yield the following {exact penalty result {for} problem~\eqref{eq:penalized-dc}.}}
{\begin{theorem}[Exact penalization]\label{cor:exact-penalty}
\color{black}
If $\tau>{8K^2\sqrt n\norm{\bA}_2^2}$, then
{problems}~\eqref{eq:penalized-dc} and \eqref{eq:kmeans-original} have the same set of
global minimizers.  %
\end{theorem}}

\section{The Proposed RADA-DC Algorithm}
In this section, we propose RADA-DC for solving a general nonsmooth
Riemannian DC optimization problem and establish its iteration complexity.
Specifically, we consider
\begin{equation}\label{eq:general-dc}
 \min_{\bX\in\Mcal}\ \left\{\Psi(\bX):=f(\bX)+h(\bX)-g(\bX)\right\},
\end{equation}
where $\Mcal$ is a Riemannian submanifold embedded in a finite-dimensional
Euclidean space $\mathcal E$, $f:\mathcal E\to\R$ is smooth but possibly
nonconvex, and $h,g:\mathcal E\to{\R}$ are closed and
convex, but possibly nonsmooth.  In particular, problem
\eqref{eq:general-dc} includes the {DC penalty formulation \eqref{eq:penalized-dc} for K-means} as a special case.

\subsection{Proposed Algorithm}

Before presenting RADA-DC, we introduce some basic objects associated with
the Riemannian manifold $\Mcal$ \cite{boumal2023intromanifolds}.
For $\bX\in\Mcal$, let $\mathrm{T}_{\bX}\Mcal$ be the tangent space {to $\Mcal$ at $\bX$} and
let $\Proj_{\mathrm{T}_{\bX}\Mcal}$ denote the orthogonal projection onto
$\mathrm{T}_{\bX}\Mcal$.  Since $\Mcal$ is endowed with the metric inherited from
$\mathcal E$, the Riemannian gradient of a smooth function $\theta$ is
$\grad\theta(\bX)=\Proj_{\mathrm{T}_{\bX}\Mcal}(\nabla\theta(\bX))$.
{Let} $\Retr_{\bX}:\mathrm{T}_{\bX}\Mcal\to\Mcal$ be a retraction {at $\bX$} satisfying
$\Retr_{\bX}({\bzero_{\bX}})=\bX${, where $\bzero_{\bX}$ is the zero element in $\mathrm{T}_{\bX}\Mcal$, and
$\left.\frac{\mathrm d}{\mathrm dt}\Retr_{\bX}(t\bxi)\right|_{t=0}=\bxi$ for every $\bxi\in\mathrm{T}_{\bX}\Mcal$.}

We next recall several standard notions from convex analysis \cite{beck2017first}.  For a proper
closed convex function $p:\mathcal E\to(-\infty,+\infty]$, its
conjugate function {is}
\begin{equation}\label{eq:conjugate}
 p^*(\bY):=\sup_{{\bU\in\mathcal E}}\{\ip{\bU}{\bY}-p(\bU)\},
\end{equation}
and its convex subdifferential is
$\partial p(\bX):=\{\bZ\in\mathcal E \mid p(\bU)\geq p(\bX)+\ip{\bZ}{\bU-\bX},\ \forall\,\bU \in \mathcal{E}\}$.
We denote $\operatorname{dom}p:=\{\bU \in \mathcal{E} \mid p(\bU)<+\infty\}$ as its domain.
For $\eta>0$, the proximal mapping of $h$ is {defined as}
\begin{equation}\label{eq:prox-h}
 \prox_{\eta h}(\bU):=\arg\min_{{\bV\in\mathcal E}}
 \left\{h(\bV)+\frac{\|\bV-\bU\|^2}{2\eta}\right\}.
\end{equation}

We are now ready to present RADA-DC.  First, by \eqref{eq:conjugate} {with $p=h^*$ and the identity $h=h^{**}$,} problem
\eqref{eq:general-dc} can be equivalently reformulated as
\begin{equation}\label{eq:minimax-F}
 \min_{\bX\in\Mcal}\max_{\bY\in\Ycal}\
 f(\bX)-g(\bX)+\ip{\bX}{\bY}-h^*(\bY),
\end{equation}
where $\Ycal\mathrel{{=}}\operatorname{dom}h^*$.
{Motivated by the RADA method in \cite{xu2024riemannian}, we first regularize the inner maximization in \eqref{eq:minimax-F} to obtain a smooth approximation of $f+h$. We then adopt the standard DC linearization strategy \cite{lethi2018dc} to replace the remaining nonsmooth concave term $-g$ with an affine upper bound, yielding a smooth subproblem in $\bX$ on $\Mcal$. The algorithm alternates between approximately solving this subproblem and updating the dual variable $\bY$.} 

{Specifically,} at the $k$-th iteration, {given the proximal center $\bY^k$,} define the regularized value function
\begin{equation}\label{eq:value-function}
 \begin{aligned}
 \varphi_k(\bX):=f(\bX)+\max_{\bY\in\Ycal}\Big\{&\ip{\bX}{\bY}-h^*(\bY)\\
 &-\frac{\lambda}{2}\|\bY\|^2
 -\frac{\beta_k}{2}\|\bY-{\bY^k}\|^2\Big\},
 \end{aligned}
\end{equation}
where $\lambda>0,\,{\beta_k\geq0}$ are regularization parameters. {For some $\rho>1$, we choose $0\leq\beta_k\leq\beta_1/k^\rho$.} {Next, for given} $\bX^k$, choose any $\bZ^k\in\partial g(\bX^k)$. The convexity of $g$ implies {that}
$-g(\bX)\leq-g(\bX^k)-\ip{\bZ^k}{\bX-\bX^k}$. {Combining this affine upper bound with the regularized value function $\varphi_k(\bX)$,} RADA-DC computes the update $\bX^{k+1}$ by approximately {solving}
\begin{equation}\label{eq:majorant}
\min_{\bX\in\Mcal} {\left\{\widehat\Phi_k(\bX):=\varphi_k(\bX)-g(\bX^k)-\ip{\bZ^k}{\bX-\bX^k}\right\}}.
\end{equation}
{By}
\cite[Section 3.1]{xu2024riemannian}{, the resulting value function} $\widehat\Phi_k$ is differentiable and its Riemannian gradient is 
\begin{equation}\label{eq:majorant-rgrad}
	\grad\widehat\Phi_k(\bX)
	=\Proj_{\mathrm{T}_{\bX}\Mcal}\big(\nabla f(\bX)+{\bY^k_\star}(\bX)-\bZ^k\big)\rev{,}
\end{equation}
\rev{where ${\bY^k_\star}(\bX)$ denotes the unique maximizer of the inner maximization problem in
\eqref{eq:value-function}, and it is given by}
\begin{equation*}
 \rev{{\bY^k_\star}(\bX)=\frac{1}{\lambda+\beta_k}\big[\bX+\beta_k{\bY^k}-\prox_{(\lambda+\beta_k)h}(\bX+\beta_k{\bY^k})\big].}
\end{equation*}
We update the variable $\bX$ by applying
$T$ Riemannian gradient descent steps to approximately
solve the subproblem \eqref{eq:majorant}{, where $T$ is a prescribed positive integer.}
Specifically, starting from $\bX_{k,1}=\bX^k$, the $t$-th inner update {with $1\leq t\leq T$} is {given by}
\begin{equation*}
 {\bX_{k,t+1}=\Retr_{\bX_{k,t}}(-\alpha_{k,t}\bD_{k,t}),}
\end{equation*}
{where $\bD_{k,t}=\grad\widehat\Phi_k(\bX_{k,t})$.}
As {in} RADA-RGD \rev{\cite[Algorithm~3]{xu2024riemannian}}, the stepsize $\alpha_{k,t}$ is selected by a backtracking line search initialized with a safeguarded Riemannian
Barzilai--Borwein (BB) trial stepsize $\zeta_{k,t}$. {The} line
search finds the smallest nonnegative integer $j_{k,t}$ such that
$\alpha_{k,t}:=\zeta_{k,t}\eta^{j_{k,t}}$ satisfies
\begin{equation}\label{eq:line-search}
 {\widehat\Phi_k(\bX_{k,t+1})-\widehat\Phi_k(\bX_{k,t})
 \leq-c_1\alpha_{k,t}\|\bD_{k,t}\|^2.}
\end{equation}
Here, $\eta,c_1\in(0,1)${.}
After $T$ inner steps, we set $\bX^{k+1}:=\bX_{k,T+1}$
{and} update the dual variable by
${\bY^{k+1}}:={\bY^k_\star}(\bX^{k+1})$.  %
The RADA-DC algorithm is formally presented in Algorithm~\ref{alg:rada-dc}.

\begin{algorithm}[t]
\caption{{RADA-DC for solving problem \eqref{eq:general-dc}}}
\label{alg:rada-dc}
\begin{algorithmic}[1]
\small
\STATE Input $\bX^1\in\Mcal$, ${\bY^1}\in\Ycal$, $\lambda>0$,
$\beta_1\geq0$, $\rho>1$, {$\eta,c_1\in(0,1)$, and}
$T\geq1$.
\FOR{$k=1,2,\ldots$}
 \STATE Choose $\bZ^k\in\partial g(\bX^k)$ and set
 $\bX_{k,1}=\bX^k$.
 \FOR{$t=1,{2,}\ldots,T$}
  \STATE Select $\alpha_{k,t}$ by the safeguarded BB-initialized
  backtracking line search so that \eqref{eq:line-search} holds.
  \STATE {Update} $\bX_{k,t+1}=\Retr_{\bX_{k,t}}(-\alpha_{k,t}\bD_{k,t})$.
 \ENDFOR
 \STATE Set $\bX^{k+1}=\bX_{k,T+1}$.
 \STATE Update ${\bY^{k+1}}={\bY^k_\star}(\bX^{k+1})$.
 \STATE Choose $0\leq\beta_{k+1}\leq\beta_1/(k+1)^\rho$.
\ENDFOR
\end{algorithmic}
\end{algorithm}
\vspace{-3pt}
\subsection{Convergence Analysis}

Following \cite[Definition~2.3]{jiang2025irpdc}, we define the {notion of an} \(\epsilon\)-RC {point} used in our analysis.

\begin{definition}\label{def:stationarity}
A point $\bX\in\Mcal$ is called an {$\epsilon$-RC} point of
\eqref{eq:general-dc} if there exist $\bW\in\mathcal E$,
$\bZ\in\partial g(\bX)$, and $\bY\in\partial h(\bW)$ such that
\begin{equation}\label{eq:stationarity}
 \|\Proj_{\mathrm{T}_{\bX}\Mcal}(\nabla f(\bX)+\bY-\bZ)\|\leq\epsilon,
 \quad
 \|\bW-\bX\|\leq\epsilon.
\end{equation}
\end{definition}
We next make the following standard assumptions on problem
\eqref{eq:general-dc}.

\begin{assumption}\label{ass:model}
The {manifold} $\Mcal$ is a compact submanifold embedded in {the} Euclidean space $\mathcal E$.  The function $f:\mathcal E\to\R$ is
continuously differentiable on {$\mathcal{E}$}, with {an} $L_f$-Lipschitz gradient {for some $L_f>0$}.  {The functions $h$ and $g$ are {closed, convex, and} Lipschitz continuous on {$\mathcal{E}$} with constants {$L_h, L_g > 0$}, respectively. Moreover, the proximal mapping of $h$ and a subgradient of $g$ at any $\bX\in\Mcal$ can be efficiently computed.}
\end{assumption}

{\begin{theorem}[{Iteration complexity}]\label{thm:complexity}
\color{black}
Given a constant $\epsilon>0$, let $\{\bX^{k}\}$ be the sequence generated by Algorithm \ref{alg:rada-dc} with $\lambda=\epsilon/(2L_h)$. Suppose that Assumption \ref{ass:model} holds. Then, there exists an integer ${\hat{k}=\mathcal{O}(\epsilon^{-3})}$ such that $\bX^{\hat{k}}$ is an {$\epsilon$-RC} point of {problem} \eqref{eq:general-dc}.
\end{theorem}}
\vspace{-9pt}
\subsection{{Customized Application}}
{In this subsection, we apply the proposed Algorithm~\ref{alg:rada-dc} to solve the DC penalty formulation~\eqref{eq:penalized-dc} for K-means.}

{For {problem~\eqref{eq:penalized-dc}}, {we have} $\Mcal=\Fcal$, $f=f_{\rm km}$, $h(\bX)=\tau\norm{\bX}_1$, and $g(\bX)=\tau\norm{\bX}_{[n]}$. Assumption~\ref{ass:model} holds with $L_f=2\norm{\bA\bA^\top}_2$, $L_h=\tau\sqrt{nK}${, and $L_g=\tau\sqrt n$.} In particular, $\Fcal$ is compact, \rev{and} $h$ and $g$ are finite-valued Lipschitz convex functions. 
{Since $h^*$ is the indicator function of the box $[-\tau,\tau]^{n\times K}$,}
$\bY^k_\star(\bX)$ is {obtained by projecting} $(\bX+\beta_k\bY^k)/(\lambda+\beta_k)$ {onto this box}. {A subgradient $\bZ\in\partial g(\bX)$ is obtained by setting $\bZ_{ij}=\tau\operatorname{sgn}(\bX_{ij})$ on the $n$ largest-magnitude entries of $\bX$ (breaking ties arbitrarily) and zero elsewhere\rev{, where $\operatorname{sgn}$ denotes the sign function with $\operatorname{sgn}(0)=0$.}} We use the {closed-form} tangent projection and \rev{the retraction based on polar decomposition} in \cite[Eq.~(3.9)]{huang2025riemannian}.}
\vspace{-3pt}
\section{Numerical Experiments}
In this section, we compare  RADA-DC  with the inexact accelerated manifold proximal-gradient method (I-AManPG) \cite{huang2025riemannian} and K-means++ \rev{(KM++)} \cite{arthur2007kmeans}. {Although RADA-DC and I-AManPG are both manifold-based methods, they are applied to different continuous reformulations: RADA-DC solves our proposed DC penalty model, which is globally exact for a sufficiently large finite penalty parameter, whereas I-AManPG solves the $\ell_1$-regularized model in \cite{huang2025riemannian}, for which such equivalence is not generally guaranteed.}
{Our experiments focus on large-$K$ settings. {We do not include the second-order method in \cite{xu2026scalable} because its $\mathcal O(nr^3)$ per-iteration cost, with a search rank $r>K$ required for strict feasibility, can be computationally demanding in these settings.}}

Each {experiment} setting is repeated {50} times with different random seeds. %
{RADA-DC and I-AManPG} share the same feasible spectral initialization in each repetition.  {For each of these two methods, the cluster centers derived from its solution initialize one complete run of Lloyd's algorithm, whose output is used as the final clustering result. For K-means++, we perform 1000 independently initialized runs in each repetition and use the solution with the lowest final objective value.} 
\begin{figure}[!t]
	\centering
	{\includegraphics[width=\columnwidth]{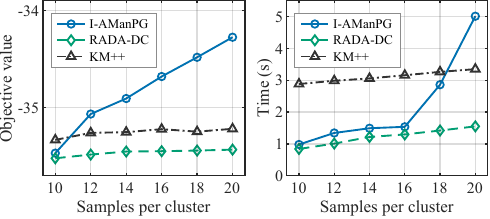}}
	\caption{{Average results over 50 runs on synthetic data with $K=40$.} %
	}
	\label{fig:synthetic}
\end{figure}
We set $\epsilon=10^{-2}$ and
{$\tau=\mu_0 K^2\sqrt n \norm{\bA}_2^2$ {with} $\mu_0=2\times10^{-6}$.}
{Since the {theoretical threshold} for exact penalization in Theorem~\ref{cor:exact-penalty} can be overly conservative in practice, we choose $\mu_0$ based on preliminary experiments and keep it fixed across all tested instances.} The RADA-DC parameters are {$T=5$,} $\beta_1=10n\sqrt{K}$, $\rho=1.5$, and $\lambda=10^{-12}$. {This choice of $\lambda$ is consistent with the theoretical parameter setting when a larger valid Lipschitz bound is used as $L_h$.} 
{I-AManPG uses a relative proximal-direction stationarity criterion with tolerance $\epsilon$ \cite{huang2025riemannian}, whereas RADA-DC stops at an $\epsilon$-RC point in the sense of Definition~\ref{def:stationarity}. Every reported run of these two methods satisfies its respective {criteria}.} For the normalized
indicator matrix $\bX$ corresponding to the final partition, we report the objective value {(``Obj.'')}
$f_{\rm km}(\bX)$, where a lower value is better.
The clustering error {(``Err.'')} is {the fraction of misclassified samples} after optimally matching the cluster indices
{with} the ground-truth labels. {For RADA-DC and I-AManPG, the} reported total time includes spectral
initialization, {nonsmooth Riemannian} optimization, construction of the initial {cluster} centers,
and {the subsequent execution of Lloyd’s algorithm.} For K-means++, the reported time includes all 1000 independent starts.
\vspace{-3pt}
\subsection{Synthetic Data}
{Motivated by the synthetic data constructions for K-means clustering in \cite{chen2019bigdata,derosa2026power}, we arrange the cluster centers so that their pairwise distances are comparable and sample data points around each center.}
{Let $\bm e_1,\,\bm e_2, \ldots,\bm e_{K-1}\in\R^{300}$ denote the first $K-1$ standard basis vectors. We place the $K$ cluster centers at
$
\{\bzero,\, \bm e_1,\,\bm e_2,\ldots,\bm e_{K-1}\}-\bm\mu$ with  $\bm\mu=\sum_{i=1}^{K-1}\bm e_i/K$\rev{, so that their mean is zero.}
We fix $K=40$ and \rev{for each} $s\in\{10,12,14,16,18,20\}$\rev{, independently draw $s$ samples} per
cluster uniformly from the ball \rev{of radius} $0.75$ \rev{centered at the corresponding cluster center}. As preprocessing, we retain the leading $K$ left singular vectors of the resulting data matrix as the input matrix $\bA$ to all methods.}

Figure~\ref{fig:synthetic} shows that RADA-DC attains
the lowest mean objective value and requires substantially less total running
time than K-means++. As $s$ increases,
the mean objective value obtained by I-AManPG increases appreciably, whereas
that obtained by RADA-DC remains nearly unchanged. The running time of RADA-DC also grows only moderately, in contrast to the pronounced increase observed for I-AManPG at larger values of $s$.%
\vspace{-3pt}
\subsection{Real-World Image Datasets}
\begin{table}[!t]
	\caption{Average results over {50} runs on ORL40 and
		Faces94, shown outside and inside parentheses, respectively.
		}
	\label{tab:orl_faces94}
	\centering
	\setlength{\tabcolsep}{1pt}
	\begin{tabular}{@{}clccc@{}}
		\toprule
		$n$ & Method & $-\mathrm{Obj.}\uparrow$ & Err. (\%)$\downarrow$ & Time (s)$\downarrow$\\
		\midrule
		{200\,(500)} & I-AManPG & {{22.35}\,({40.30})} & {{33.24}\,({16.14})} & {0.83\,({3.65})}\\
		& RADA-DC   & {\textBF{{22.60}}\,(\textBF{{41.18}})} & {\textBF{{31.95}}\,(\textBF{{11.77}})} & {\textBF{{0.62}}\,(\textBF{{1.49}})}\\
		& KM++      & {{21.75}\,({39.97})} & {{37.84}\,({15.98})} & {{2.75}\,({4.00})}\\
		\addlinespace[1pt]
		{240\,(550)} & I-AManPG & {{21.87}\,({40.09})} & {{33.18}\,({17.03})} & {{0.91}\,({4.97})}\\
		& RADA-DC   & {\textBF{{22.00}}\,(\textBF{{41.21}})} & {\textBF{{32.48}}\,(\textBF{{11.99}})} & {\textBF{{0.67}}\,(\textBF{{1.68}})}\\
		& KM++      & {{21.22}\,({40.08})} & {{36.84}\,({15.29})} & {{2.85}\,({4.03})}\\
		\addlinespace[1pt]
		{280\,(600)} & I-AManPG & {{21.50}\,({39.70})} & {{33.31}\,({18.81})} & {{0.85}\,({5.41})}\\
		& RADA-DC   & {\textBF{21.60}\,(\textBF{41.17})} & {\textBF{{32.56}}\,(\textBF{{11.62}})} & {\textBF{{0.76}}\,(\textBF{{1.80}})}\\
		& KM++      & {{20.81}\,({40.08})} & {{36.07}\,({15.10})} & {{3.01}\,({4.25})}\\
		\addlinespace[1pt]
		{320\,(650)} & I-AManPG & {{21.23}\,({39.40})} & {{33.56}\,({19.34})} & {{0.88}\,({5.86})}\\
		& RADA-DC   & {\textBF{{21.37}}\,(\textBF{{41.16}})} & {\textBF{{32.49}}\,(\textBF{{11.47}})} & {\textBF{0.82}\,(\textBF{{1.98}})}\\
		& KM++      & {20.60\,({40.05})} & {{35.56}\,({15.28})} & {{3.22}\,({4.50})}\\
		\addlinespace[1pt]
		{360\,(700)} & I-AManPG & {{21.09}\,(39.29)} & {{33.87}\,({19.95})} & {\textBF{0.89}\,({5.61})}\\
		& RADA-DC   & {\textBF{{21.18}}\,(\textBF{{41.17}})} & {\textBF{{32.98}}\,(\textBF{{11.52}})} & {{0.90}\,(\textBF{{2.03}})}\\
		& KM++      & {{20.39}\,({40.08})} & {{35.40}\,({15.78})} & {3.32\,({4.25})}\\
		\bottomrule
	\end{tabular}
\end{table}

{For a more comprehensive evaluation,} we {test} the three methods on two real-world datasets,
ORL40 and Faces94, where each class corresponds to one individual. ORL40 contains $K=40$ classes with ten $92\times112$ {grayscale} images per class
\cite{samaria1994parameterisation}. {Using} all 40 classes,
we construct instances with $n=200,240,\ldots,360$ \rev{in each run} by \rev{randomly} sampling $n/40$ images per class. \rev{For Faces94~\cite{arora2024rtlbp,spacekfaces94}, we select $K=50$ classes  from a pool of 152 classes, each containing 20 color images. For each $n\in\{500,550,\ldots,700\}$, we construct an instance in each run by randomly sampling $n/50$ images per selected class.} 
For each instance, the selected images are converted to {grayscale} when necessary, vectorized, and standardized. We {use} the leading
$K$ left singular vectors of the standardized data matrix {as} the input matrix {$\bA$ for all methods}. 

Table~\ref{tab:orl_faces94} shows that RADA-DC attains the lowest mean objective value and clustering error on both datasets. {It also has the shortest mean running time in all but the ORL40 setting with $n=360$.} 

{In summary, we developed a Riemannian DC approach to K-means clustering, establishing the exactness of the penalty formulation for sufficiently large finite penalty parameters and an $\mathcal O(\epsilon^{-3})$ iteration complexity guarantee for RADA-DC. Our numerical results demonstrate that this approach achieves lower objective values on average than the tested baselines at competitive computational cost in the {large-$K$ settings}.}

\bibliographystyle{IEEEtran-rev}
{\bibliography{reference_rada_dc}}
\end{document}